\documentclass[journal]{IEEEtran}

\usepackage{times}
\usepackage{epsfig}
\usepackage{graphicx}
\usepackage{amsmath}
\usepackage{amssymb}
\usepackage{booktabs}

\usepackage{xcolor}
\usepackage{enumitem}
\usepackage{lipsum}
\usepackage{multirow}

\usepackage[font=small, skip=0.3pt]{caption}
\usepackage[ruled,vlined,linesnumbered]{algorithm2e}
\usepackage[multiple]{footmisc}
\usepackage{float}

\usepackage{tcolorbox}
\usepackage{colortbl}
\definecolor{lavender}{rgb}{0.9, 0.9, 0.98}
\definecolor{magenta}{rgb}{0.96, 0.0, 0.8}

\usepackage{pifont}
\newcommand{\cmark}{\ding{51}}%
\newcommand{\xmark}{\ding{55}}%
\newcommand{\ie}{{\it i.e.}}
\newcommand{\eg}{{\it e.g.}}

\usepackage[breaklinks=true,bookmarks=false]{hyperref}

\begin{document}

\title{Hard to Track Objects with Irregular Motions and Similar Appearances? \\Make It Easier by Buffering the Matching Space }

\author{
Fan Yang, Shigeyuki Odashima, Shoichi Masui, Shan Jiang\\
Fujitsu Research, Japan\\
{\tt\small contact: hongheyangfan@gmail.com; fan.yang@fujitsu.com}
}
\maketitle
\thispagestyle{empty}

\begin{abstract}

   We propose a Cascaded Buffered IoU (C-BIoU) tracker to track multiple objects that have irregular motions and indistinguishable appearances. When appearance features are unreliable and geometric features are confused by irregular motions, applying conventional Multiple Object Tracking (MOT) methods may generate unsatisfactory results. To address this issue, our C-BIoU tracker adds buffers to expand the matching space of detections and tracks, which mitigates the effect of irregular motions in two aspects: one is to directly match identical but non-overlapping detections and tracks in adjacent frames, and the other is to compensate for the motion estimation bias in the matching space. In addition, to reduce the risk of overexpansion of the matching space, cascaded matching is employed: first matching alive tracks and detections with a small buffer, and then matching unmatched tracks and detections with a large buffer. Despite its simplicity, our C-BIoU tracker works surprisingly well and achieves state-of-the-art results on MOT datasets that focus on irregular motions and indistinguishable appearances. Moreover, the C-BIoU tracker is the dominant component for our $2^{nd}$ place solution in the CVPR'22 SoccerNet MOT and ECCV'22 MOTComplex DanceTrack challenges. Finally, we analyze the limitation of our C-BIoU tracker in ablation studies and discuss its application scope.

\end{abstract}


\section{Introduction}

\begin{figure}[h!]
       \centering
       \includegraphics[width=\linewidth]{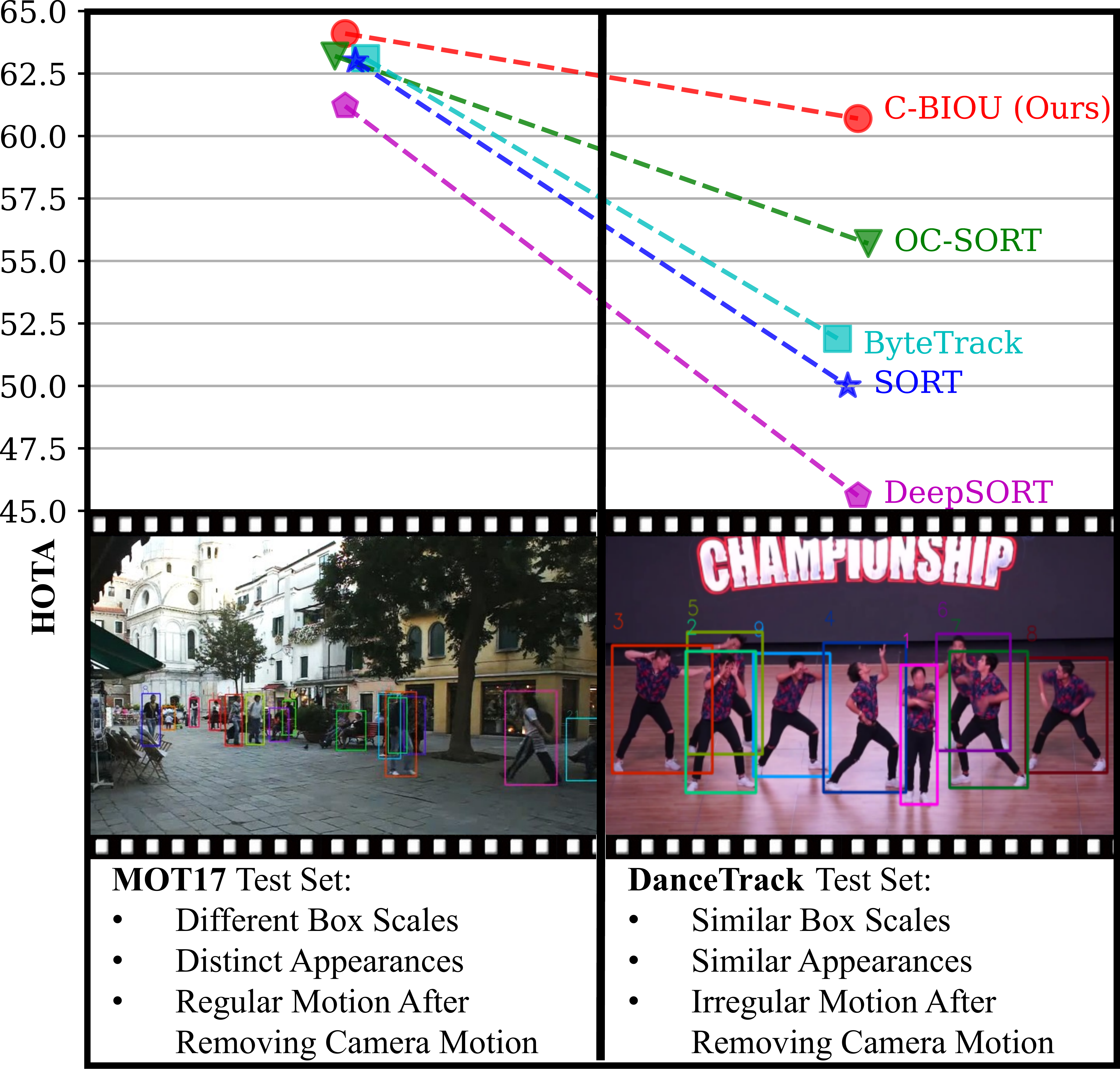}
       \caption{\textbf{Tracking performance on the test sets of MOT17~\cite{MOT16} and DanceTrack~\cite{sun2022dancetrack}.} For a fair comparison, all methods are \textit{online approaches} and use detections generated by YOLOX-X~\cite{ge2021yolox}. On the MOT17, our method has a similar HOTA score to other methods, whereas on the DanceTrack, our method increases the HOTA score by a remarkable margin compared to DeepSORT~\cite{DeepSORT}, SORT~\cite{SORT}, ByteTrack~\cite{bytetrack}, and OC-SORT~\cite{cao2022observation}.}
       \label{fig:demo}
\end{figure}

\begin{figure}[th!]
   \centering
   \includegraphics[width=\linewidth]{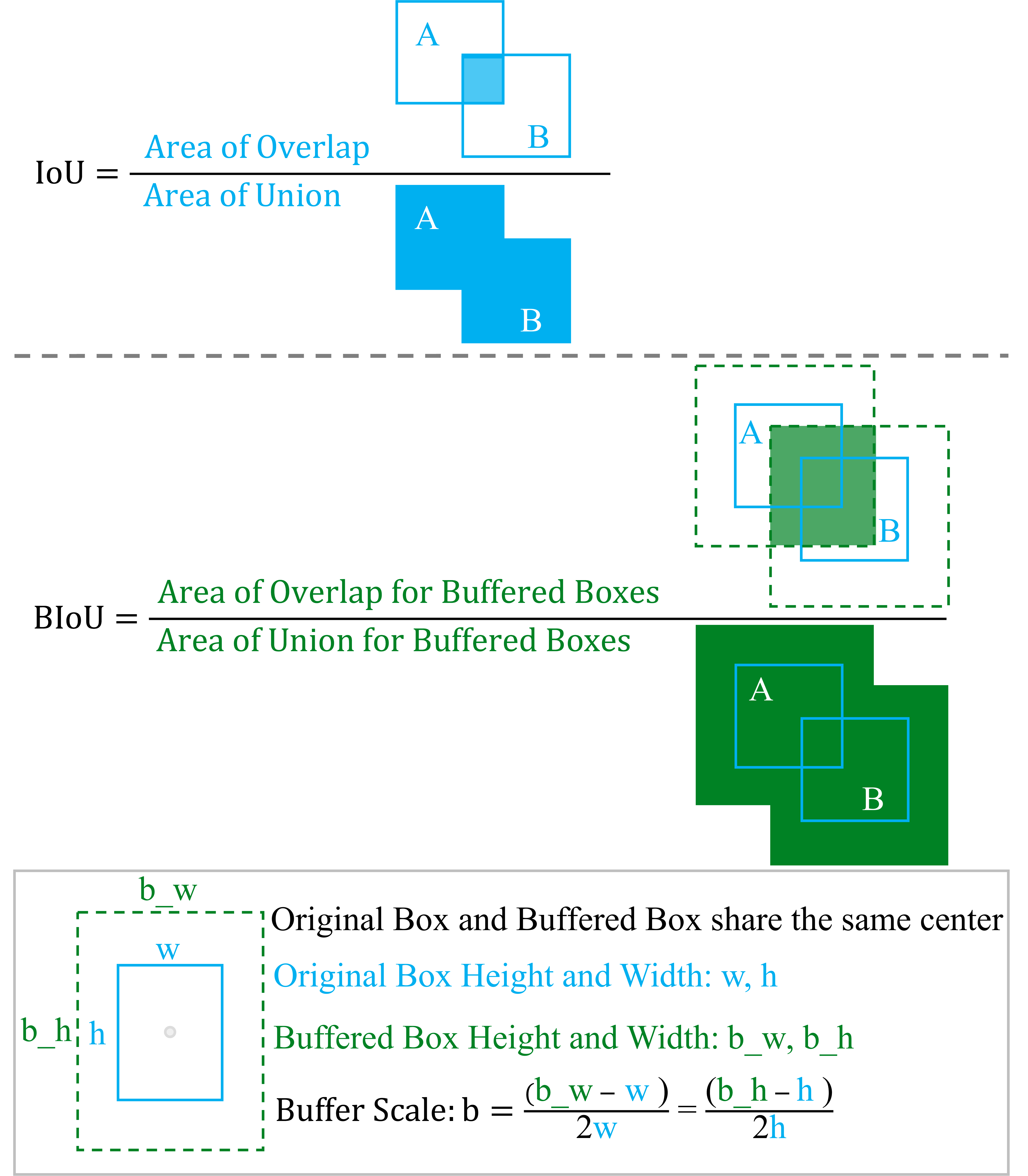}
   \caption{\textbf{Illustration of how Buffered IoU (BIoU) is calculated}. Our BIoU adds a buffer that is proportional to the original bounding box. It does not change the location center, scale ratio, and shape of the original bounding boxes but expands the original matching space.}
   \label{fig:BIoU}
\end{figure}

Multiple Object Tracking (MOT) is widely applied to identify the trajectory of each object in sequential data (\eg, videos). It offers important information for real-world applications which include but are not limited to autonomous driving~\cite{KITTI}, sports and dance analysis~\cite{sun2022dancetrack,cioppa2022soccernet}, and animal surveys~\cite{bai2021gmot,lauer2022multi}. 

Although MOT studies have been greatly developed~\cite{SORT,DeepSORT,JDE,FairMOT,bytetrack, Li_2022_CVPR},  a new challenge has recently attracted attention: unlike conventional MOT tasks that contain objects with distinct appearances and regular motions, MOT tasks that cover animals, group dancers, and sports players, may have indistinguishable appearances and irregular motions, which could cause existing MOT methods to fail. In particular, as shown in Fig.~\ref{fig:demo}, several MOT methods~\cite{SORT, DeepSORT, bytetrack, cao2022observation} that perform well on MOT17~\cite{MOT16}, may experience a significant performance drop on the DanceTrack~\cite{sun2022dancetrack}.

Why does the HOTA score drop significantly on the DanceTrack? \textbf{We presume that tracking failures are caused by two reasons: 
(i) The detections and tracks of identical objects do not overlap between adjacent frames (\eg, due to the fast movement) and thus the tracking fails; (ii) After track initialization, unmatched tracks (\eg, occluded objects) continue to update their geometric features for multiple frames, however, if their motion estimations are inaccurate (\eg, due to a sudden acceleration or turning), they miss the matching opportunity when corresponding detections are available in subsequent  frames.} When the appearance of objects can be distinguished, appearance features could be employed to alleviate issues (i) and (ii), by matching cross-frame detections based on their appearance similarities. Nonetheless, when irregular motions are accompanied by indistinguishable appearances, most existing MOT solutions may not be able to perform a dependable tracking, so a new solution is desirable.

\begin{figure}[t!]
   \centering
   \includegraphics[width=0.6\linewidth]{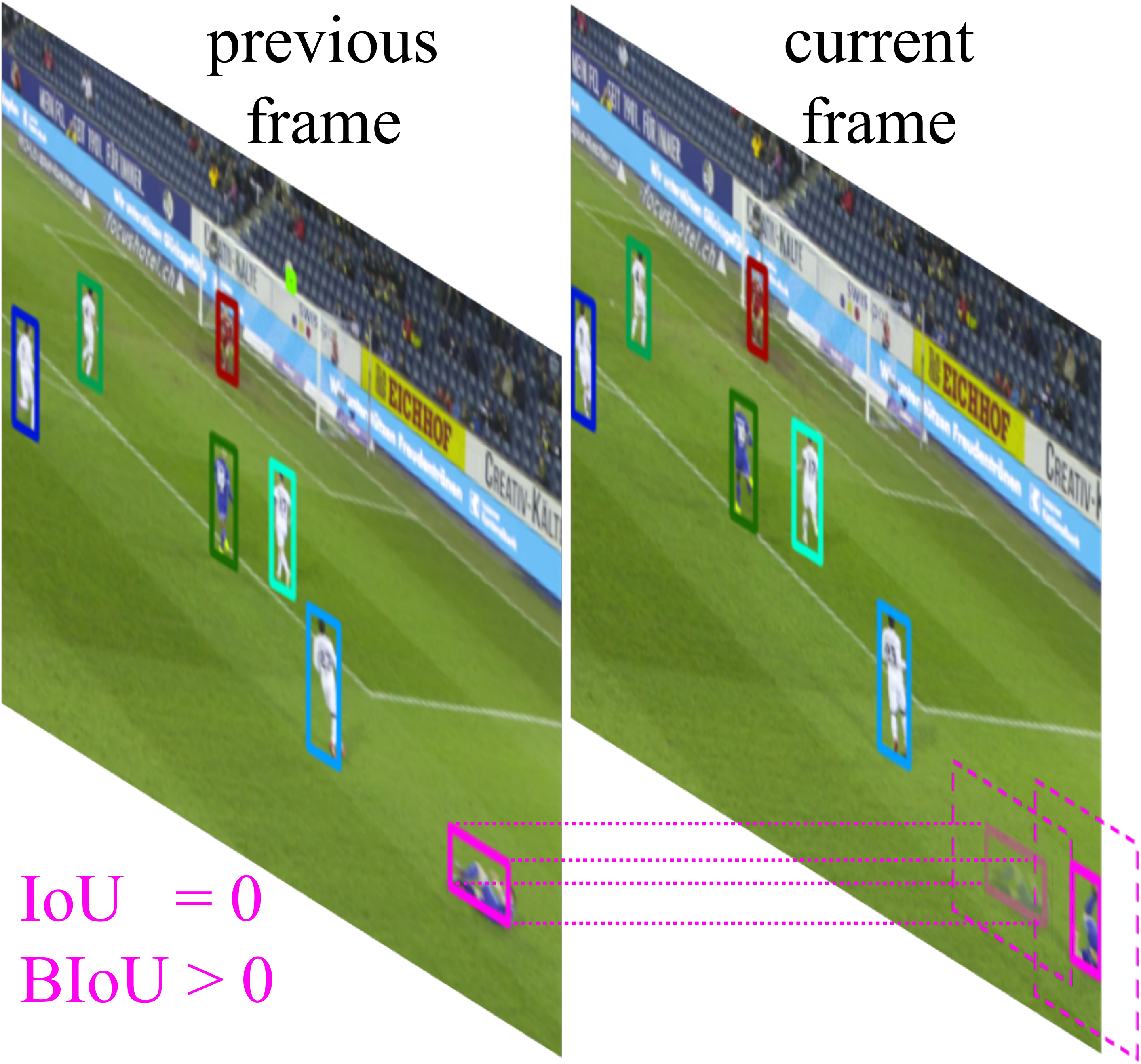}
   \caption{\textbf{An illustration of BIoU forms better cross-frame geometric consistency than IoU}. The bounding box of an identical object shares the same color. The \textcolor{magenta}{magenta object} has no overlapping detections between adjacent frames. Whether this is caused by the fast movement or incorrect motion estimation, our BIoU expands the matching space to reduce the miss matching.}
   \label{fig:BIoU_demo}
\end{figure}

In this study, we propose a Cascaded-Buffered Intersection over Union (C-BIoU) tracker to track multiple objects that have irregular motions and indistinguishable appearances. Our BIoU (Fig.~\ref{fig:BIoU}) is applied to alleviate issues (i) and (ii). Unlike the IoU, which only forms spatiotemporal similarities between overlapping detections and tracks, our BIoU constructs spatiotemporal similarities for originally non-overlapping detections and tracks if they are within the range of the buffers (Fig.~\ref{fig:BIoU_demo}). Because the buffers are proportional to the original detections and tracks, the BIoU does not change their location centers, scale ratios, and shapes but expands their matching space. With these properties, \textbf{our BIoU mitigates the effect of irregular motions in two aspects: one is to directly match identical but non-overlapping detections and tracks in adjacent frames, and the other is to compensate for the motion estimation bias in the matching space.} Additionally, to reduce the risk of matching space overexpansion, we incorporate the BIoU into a cascaded matching scheme: first, alive tracks and detections are matched using a small buffer, and then, unmatched tracks and detections are matched again using a large buffer. To this end, our C-BIoU tracker could relieve mismatching caused by irregular motions and improve the tracking performance.

We report promising results on a variety of MOT datasets~\cite{cioppa2022soccernet,deliege2021soccernet, bai2021gmot, sun2022dancetrack} that focus on irregular motions and indistinguishable appearances. Compared with other strong MOT methods (\eg, OC-SORT~\cite{cao2022observation}), our C-BIoU tracker greatly improves the tracking performance, ranging from $2.6$ to $7.2$ in terms of the HOTA score~\cite{luiten2020IJCV}. Moreover, the C-BIoU tracker is the dominant component for our $2^{nd}$ place solution in the CVPR'22 SoccerNet MOT~\cite{SoccerNet_tracking_2022} and  ECCV'22 MOTComplex DanceTrack challenges. Finally, we analyze the limitation of our C-BIoU tracker in ablation studies and discuss its application scopes.

\section{Related Works}

\subsection{Appearance Consistency and Geometric Consistency in MOT}

In MOT studies, appearance consistency and geometric consistency are two critical assumptions used for associating cross-frame detections. In general, the previous appearance of an identical object should be similar to its current appearance (\ie, appearance consistency), and its previous location and shape added to its estimated motion should be approximate to its current location and shape (\ie, geometric consistency).

In recent works, leveraging the appearance feature for MOT has achieved great success in conventional MOT datasets (\eg, MOT17~\cite{MOT16}). In particular, after transformers~\cite{vaswani2017attention} have been introduced to MOT studies~\cite{Transtrack, MOTR, Trackformer}, the appearance similarity between cross-frame detections can be measured in a highly accurate manner, which leads to a good tracking performance. Nevertheless, the DanceTrack~\cite{sun2022dancetrack} study conducted experiments to demonstrate that appearance is not always reliable when tracking targets share a similar appearance. Other MOT datasets, such as SoccerNet~\cite{cioppa2022soccernet,deliege2021soccernet} and GMOT-40~\cite{bai2021gmot}, also reveal the challenge of real-world MOT tasks: tracking targets may look similar, which could fail MOT methods (\eg, ~\cite{bytetrack}) that achieved a state-of-the-art performance on conventional MOT datasets (\eg, MOT17~\cite{MOT16}).

Geometric matching can reduce the ambiguity caused by indistinguishable appearances. In general, the IoU is commonly used to measure geometric consistency~\cite{IoUTracker2017, SORT,DeepSORT,JDE,FairMOT,bytetrack,cao2022observation}. The IoU scores, between detections and track predictions, are used to represent their cross-frame affinity. To estimate motions, Neural Networks~\cite{MilanRD0S17} and Bayesian filters~\cite{bar1990tracking, gabriel2003state} have been typically applied. While most MOT methods~\cite{SORT,DeepSORT,JDE,FairMOT,bytetrack} apply the Kalman filter~\cite{kalman1960new} due to its simplicity, OC-SORT~\cite{cao2022observation} has enhanced the Kalman filter to handle crowded and occluded scenes. In real practice, however, motion modeling may not always be accurate. In some scenarios, for instance, soccer players and dancers may make irregular motions, which
cause the motion estimation model to fail. Additionally, for a non-stationary camera, although image registration~\cite{Cioppa_2021_CVPR} can be used to calibrate camera movements, it is time-consuming, and the accuracy cannot be guaranteed. To alleviate these problems, we introduce a new geometric consistency measurement solution.


\subsection{Geometric Consistency Measurement}

When irregular motions are given, it is difficult to initialize and estimate the motion correctly, which may result in identical objects with no overlapping geometric features in adjacent frames. Because the IoU produces the same value of $0$ for all non-overlapping geometric features (\ie, bounding boxes), using the IoU for geometric consistency measurement may fail tracking initialization and ongoing tracking. Thus, we propose a BIoU to expand the original matching space to measure the geometric consistency, which is robust to fast motions and motion estimation bias. Unlike the searching window of a previous work~\cite{yang2022tackling}, which applies the expanded bounding box as a spatial constraint, our BIoU takes the expanded bounding box as a matching feature. To some extent, using the GIoU~\cite{giou2019} and DIoU~\cite{zheng2020distance} mitigates the same issue as our BIoU does, but we verified that our BIoU may generate better results under the same conditions (Sec.~\ref{sec:ablation}).

\subsection{Cascaded Matching}
\begin{figure*}[h!]
   \centering
   \includegraphics[width=\linewidth]{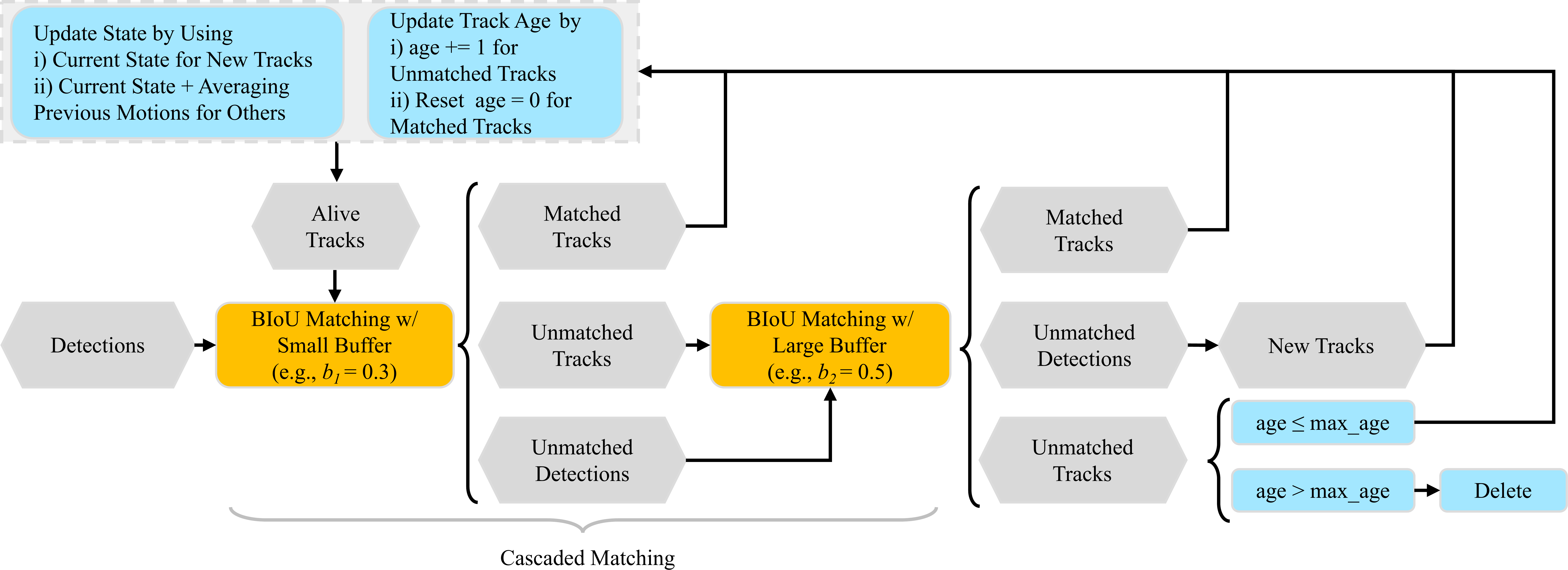}
   \caption{\textbf{Framework of Our C-BIoU Tracker.} Which initializes tracks from unmatched detections, applies the alive tracks to match new detections, and terminates a track when it has not been matched for a given amount of frames (\ie, $max\_age$). Two BIoUs, which respectively equip small and large buffers, are grouped into a cascaded matching. First, we match alive tracks and detections with the BIoU that has a small buffer (\ie, $b_1$). Then, we continue to match unmatched tracks and detections with the BIoU that has a large buffer (\ie, $b_2$). For the motion estimation, we simply average the speeds of recent frames to quickly respond to unpredictable motion changes. }
   \label{fig:C-BIoU_framework}
\end{figure*}

After obtaining the cross-frame consistency measurements, matching (\ie, data association) is applied to correspond cross-frame detections. In addition to the cross-frame consistency, we can also employ other strategies to optimize the matching process. Cascaded matching is a commonly used approach in MOT studies: matching the confident and easy samples first, followed by ambiguous and difficult samples. For example, ByteTrack\cite{bytetrack} matches confident detections earlier than unconfident detections, while DeepSORT~\cite{DeepSORT} applies data association to recently matched tracks before earlier matched tracks.
Since our BIoU changes the matching space, using a large buffer scale takes a higher risk of overexpansion than using a small buffer scale. We therefore integrate the BIoU and cascaded matching in our tracker (Fig.~\ref{fig:C-BIoU_framework}). We first match alive tracks and detections with a small buffer, and then match unmatched tracks and retained detections with a large buffer.

\section{C-BIoU Tracker}

The architecture of our Cascaded-Buffered IoU (C-BIoU) tracker is illustrated in Fig.~\ref{fig:C-BIoU_framework}. It is specifically designed to track multiple objects that have indistinguishable appearances and irregular motions. We inherit part of the track management from SORT~\cite
{SORT} and propose our C-BIoU for geometric consistency measurement.

\subsection{Tracking Pipeline}
Our tracking pipeline follows the tracking-by-detection paradigm---the object detector and MOT framework are separately designed.  Given a video, we apply the off-the-shelf object detector (\eg, YOLOX~\cite{ge2021yolox}) to generate bounding boxes at each frame. Our C-BIoU tracker then takes those bounding boxes as inputs to produce tracking results. Such a pipeline provides great flexibility to apply our C-BIoU tracker on arbitrary detections. In our experiments (Sec.~\ref{sec:main_results}), we also show that the similar pipeline~\cite{SORT, DeepSORT, bytetrack, cao2022observation} yielded strong results on our target datasets.

\subsection{Buffered IoU}
The Buffered IoU (BIoU) is our main contribution in this work. As shown in Fig.~\ref{fig:BIoU}, the BIoU simply adds buffers that are proportional to the original detections and tracks for calculating the IoU. Our BIoU retains the same location centers, scale ratios, and shapes of the original detections and tracks, but it expands the matching space to measure the geometric consistency. Let $\boldsymbol{o} = (x, y, w, h)$ denote an original detection and $(x,y,w,h)$ be the top-left coordinate, width, and height of the detection, respectively. Suppose that the buffer scale is $b$, we have the buffered detection as $\boldsymbol{o}_{b} = ( x-bw,y-bh, w+2bw, h+2bh )$. To approach our cascaded matching, we apply grid research~\cite{bergstra2012random} to find the best combination of two buffer scales $b_{1}$ and $b_{2}$ on the training set, and then apply them to the validation set and test set. Since we have $b_{1}<b_{2}$, when we search for the combination of $b_{1}$ and $b_{2}$ within a certain range, the number of combinations is limited. Considering that the speed of our C-BIoU is fast (Table~\ref{table:inference_speed}), the grid search takes an acceptable time.

\subsection{Simple Motion Estimation}

Unlike most MOT methods~\cite{SORT,DeepSORT,JDE,FairMOT,bytetrack} that apply the Kalman filter ~\cite{kalman1960new} for state estimation, we simply average motions of recent frames to quickly respond to unpredictable motion changes. 
At frame $t$, suppose that a track has matched detections for more than $n$ frames, after $\Delta$ unmatched frames, its track state $\boldsymbol{s}$ can be represented as $\boldsymbol{s}^{t+\Delta} = \boldsymbol{s}^{t} + \frac{\Delta}{n} \sum^{t}_{i=t-n+1} (\boldsymbol{o}_s^{i}- \boldsymbol{o}_s^{i-1})$, where $\boldsymbol{o}_s = (x, y, x+w, y+h)$. The matched detections between frame $t-n$ to $t$ are used to calculate motions and the average motion is applied to update the track state. We set $2\leq n \leq 5$ by default in our experiments. The IoU score of buffered $\boldsymbol{s}^{t+\Delta}_{b}$ and $\boldsymbol{o}^{i+\Delta}_{b}$ is used for data association at the frame $t+\Delta$. Due to the simplicity of our approach, the overall tracking speed is increased for our C-BIoU tracker (Table~\ref{table:inference_speed}).

\subsection{Track Management}
In an MOT framework, the function of track management is to decide how and when to initialize, update and terminate a track. We design our track management based on the mainstream solution introduced by SORT~\cite{SORT}, which is also widely applied in other well-known MOT methods~\cite{DeepSORT, FairMOT, JDE, bytetrack, cao2022observation}. 

For the first frame, we initialize all detections as new tracks. In each track, the corresponding detection is recorded in a memory. Without using the appearance information, a track may need at least two tracked frames to initialize its motion estimation. For a new track, therefore, we do not predict its motion; instead, we directly assign its recorded bounding box as its current track state. As new tracks have an age of $0$, they are all alive tracks and can be used to match detections. For the next frame, we apply the BIoU with a small buffer scale $b_{1}$ to calculate the geometric affinity between detections and alive tracks. Based on the geometric affinity, linear assignment (\eg, Hungarian algorithm~\cite{kuhn1955hungarian}) is applied to associate tracks and detections. 

After the first matching, some tracks and detections could be unmatched.
Besides the newly appeared and disappeared objects, we assume that some objects may have an inconsistency between their detections and states of the track. This inconsistency could be caused by large irregular motions. To alleviate this issue, we apply BIoU with a large buffer scale $b_{2}$ for the second matching. The first and second BIoU matching form a cascaded matching. After the second matching, we create new tracks from the unmatched detections and terminate a track when it has not been matched for a given amount of frames (\ie, $max\_age$). We update the state of a track by adding estimated motions to its current state.  Meanwhile, we also update the age of tracks. We increase the age for the unmatched tracks and reset the age to $0$ for matched tracks. This age will be compared with the threshold $max\_age$ to determine whether a track should be terminated. We repeat this progress until all frames are processed.

Note that, we only propose a simple prototype to show how to use our C-BIoU in MOT. Depending on the needs, other MOT modules can be integrated with our C-BIoU to build a more powerful MOT framework.

\begin{table*}[h!]
   \begin{minipage}[t]{.65\textwidth}
      \caption{\textbf{Results on the test sets of MOT17~\cite{MOT16} and DanceTrack~\cite{sun2022dancetrack}.} For a fair comparison, methods in the bottom block use detections generated by YOLOX-X~\cite{ge2021yolox}. On MOT17, our method has a similar HOTA score to other methods, whereas, on the DanceTrack, our method increases the HOTA score with a remarkable margin. }
      \centering
      \setlength{\tabcolsep}{0.8pt}
      \footnotesize
      \begin{tabular}{l c c c c c  c c c c c }
         \toprule
         \multirow{2}{*}{Tracker} & \multicolumn{5}{c}{MOT17 Test Set}  & \multicolumn{5}{c}{DanceTrack Test Set} \\
         \cmidrule(lr){2-6}\cmidrule(lr){7-11}
         & HOTA$\uparrow$ & DetA$\uparrow$ & AssA$\uparrow$ & MOTA$\uparrow$ & IDF1$\uparrow$~~ 
         & ~~HOTA$\uparrow$ & DetA$\uparrow$ & AssA$\uparrow$ & MOTA$\uparrow$ & IDF1$\uparrow$\\
         \hline
         \hline
         \multicolumn{11}{l}{Using Other Detections }   \\
         FairMOT~\cite{FairMOT} & 59.3 & 60.9 & 58.0 & 73.7 & 72.3 & 39.7 & 66.7 & 23.8 & 82.2 & 40.8 \\
         QDTrack~\cite{quasidense} & 53.9 & 55.6 & 52.7 & 68.7 & 66.3& 45.7 & 72.1 & 29.2 & 83.0 & 44.8\\
         TransTrack~\cite{Transtrack} &  54.1 & 61.6 & 47.9 & 75.2 & 63.5 & 45.5 & 75.9 & 27.5 & 88.4 & 45.2\\
         MOTR~\cite{MOTR} & 57.2 & 58.9 & 55.8 & 71.9 & 68.4 & 54.2 & 73.5 & 40.2	& 79.7 & 51.5 \\
         GTR~\cite{zhou2022global} & 59.1 & 61.6 & 57.0 & 75.3 & 71.5 & 48.0 & 72.5 & 31.9 & 84.7 &  50.3\\
         \hline
         \multicolumn{11}{l}{Using Detections Generated by YOLOX-x~\cite{ge2021yolox} with Input Size of [800, 1440]}   \\
         DeepSORT~\cite{DeepSORT} &61.2  &63.1 &59.7 &78.0  & 74.5 &45.6 &71.0 &29.7 &87.8 &47.9\\
         SORT~\cite{SORT} &63.0 & 64.2 & 62.2 & 80.1 & 78.2 &50.0 &75.5 & 33.2 & 90.4 & 52.0\\
         ByteTrack~\cite{bytetrack}& 63.1 & 64.5 & 62.0 & 80.3 & 77.3 &  51.9 & 80.1 & 33.8 & 90.9 & 52.0\\
         OC-SORT~\cite{cao2022observation} & 63.2 & 63.2 & 63.2 & 78.0 & 77.5 & 55.7 & \textbf{81.7} & 38.3 & \textbf{92.0} & 54.6\\
         \textbf{C-BIoU Tracker} & \textcolor{magenta}{\textbf{64.1}} & \textbf{64.8} & \textbf{63.7} & \textbf{81.1} & \textbf{79.7} & \textcolor{magenta}{\textbf{60.6}}  & 81.3  & \textbf{45.4}  & 91.6 & \textbf{61.6} \\
      \bottomrule
     \end{tabular}
     \label{table:mot17_dancetrack_test}
   \end{minipage}
   \hfill
   \begin{minipage}[t]{.33\textwidth}
      \caption{\textbf{Comparison of the tracking inference speed (w/o the detection part) using an Intel Xeon Silver 4216 CPU}. The unit is FPS (Frames Per Second). Because our C-BIoU utilizes the average speed of recent frames other than the Kalman filter for its motion estimation, it is faster than other trackers. Note that, the speed of tracker is proportional to the number of tracking objects, and when the number of objects increases, the speed of the tracker drops.}
      \centering
      \setlength{\tabcolsep}{3pt}
      \footnotesize
      \begin{tabular}{ l cc}
      \toprule
      Tracker & MOT17 & DanceTrack \\
      \midrule
      SORT~\cite{SORT} &144  &271 \\
      ByteTrack~\cite{bytetrack} & 118 & 207 \\
      OC-SORT~\cite{cao2022observation} & 185 & 341  \\
      \textbf{C-BIoU Tracker} & \textbf{361}  & \textbf{680}  \\
      \bottomrule
     \end{tabular}
     \label{table:inference_speed}
   \end{minipage}
 \end{table*}


\section{Experiments}

Our experiments consist of three parts. In Sec.~\ref{sec:setting}, we present the details of our experimental dataset and evaluation metrics. Then, in Sec.~\ref{sec:main_results}, we demonstrate the effectiveness of our C-BIoU tracker by comparing its performance to state-of-the-art methods on four MOT datasets. Next, in Sec.~\ref{sec:ablation}, we perform ablation studies to investigate (1) how our BIoU, cascaded matching, and motion modeling contribute to our final results; (2) how our dominant parameters, as the buffer scales, affect the tracking performance; and (3) how detection noise influences our C-BIoU tracker and the corresponding limitation of our C-BIoU tracker.

\subsection{Dataset and Evaluation Metrics}
\label{sec:setting}
\noindent \textbf{Datasets.}
Four public MOT datasets are used in our experiments. MOT17~\cite{MOT16} covers conventional tracking scenes: most tracking targets may have distinguishable appearances, and their motions could be regular after removing camera motions. DanceTrack~\cite{sun2022dancetrack}, SoccerNet~\cite{cioppa2022soccernet,deliege2021soccernet}, and GMOT-40~\cite{bai2021gmot} introduce another kind of realistic tracking scenario, where tacking targets share a similar texture and have irregular motions (even after removing the camera motion). Besides, compared to MOT17, more frames are included in DanceTrack, SoccerNet, and GMOT-40, which helps us make a comprehensive analysis.

\noindent \textbf{Evaluation Metrics.}
Although MOTA~\cite{bernardin2008evaluating} used to be a dominant metric for the MOT evaluation, it may favor detection over association performance. To alleviate the limitation of MOTA, the HOTA metric~\cite{luiten2020IJCV} was proposed to provide a better trade-off between detection and association performance, and thus, it is the dominant metric for recent MOT evaluations. In our experiments, we select HOTA metrics (\ie, HOTA, DetA and AssA)~\cite{luiten2020IJCV}, CLEAR metrics (\ie, MOTA)~\cite{bernardin2008evaluating} and Identity metrics (\ie, IDF1)~\cite{ristani2016performance} to evaluate the tracking results from various perspectives. Among them, the HOTA score is our dominant metric.

\noindent \textbf{Evaluation Approaches.} To evaluate the test sets of MOT17 and DanceTrack, we submit the result to their official evaluation servers
to obtain the evaluation feedback. Meanwhile, we utilize the ground truth of the DanceTrack validation set, SoccerNet test set, and GMOT-40 test set to perform evaluations with the TrackEval~\cite{luiten2020IJCV} evaluation script. In our experiments, we apply the default data splitting for DanceTrack, SoccerNet, and GMOT-40.

\subsection{Main Results}
\label{sec:main_results}


\begin{table}[t]
   \caption{\textbf{Comparisons on the DanceTrack validation set~\cite{sun2022dancetrack}, SoccerNet test set~\cite{cioppa2022soccernet,deliege2021soccernet}, and GMOT-40 test set~\cite{bai2021gmot}}. Where ``App.''  and ``Mo.'' represent the appearance feature and motion estimation, respectively. }
   
   \centering
   \setlength{\tabcolsep}{0.5pt}
   \footnotesize
   \begin{tabular}{ l ccccc}
   \toprule
   Tracker & HOTA$\uparrow$ & DetA$\uparrow$ & AssA$\uparrow$ & MOTA$\uparrow$ & IDF1$\uparrow$\\
   \midrule
   \hline
   \multicolumn{6}{l}{DanceTrack Validation Set~\cite{sun2022dancetrack}. Using Oracle Detections. }   \\
   DanceTrack (IoU) \cite{sun2022dancetrack} & 72.8 & \textbf{98.9} & 53.6 & 98.7 & 63.5\\  
   DanceTrack (IoU+Mo.) \cite{sun2022dancetrack}  & 69.4 & 87.9 & 54.8 & 99.4 & 71.3\\ 
   DanceTrack (App.) \cite{sun2022dancetrack} &  59.7 & 82.5 & 43.2 & 97.2 & 60.5\\ 
   DanceTrack (IoU+Mo.+App.) \cite{sun2022dancetrack} &   68.0 & 97.7 & 47.4 & 97.9 & 58.7\\ 
   DeepSORT~\cite{DeepSORT}  & 66.8 & 86.1 &  51.8 & 97.4  & 68.3\\
   SORT~\cite{SORT}  &67.6 & 86.6 & 52.8 & 98.1 & 69.6\\
   OC-SORT~\cite{cao2022observation}  & 79.1& 97.7 &  64.0 & \textbf{99.6}  & 76.1\\
   \textbf{C-BIoU Tracker}  &  \textcolor{magenta}{\textbf{81.7}} & 97.6 & \textbf{68.4} & 99.3 & \textbf{80.5}\\
   \hline
   \multicolumn{6}{l}{SoccerNet Test Set~\cite{cioppa2022soccernet,deliege2021soccernet}. Using Oracle Detections. }   \\
   ByteTrack~\cite{bytetrack} (reported by \cite{cioppa2022soccernet}) & 71.5 &84.3 &60.7 & 94.6 & -\\
   DeepSORT~\cite{DeepSORT} (reported by \cite{cioppa2022soccernet}) & 69.6 &82.6 &58.7 &94.8 & -\\
   SORT~\cite{SORT}  &74.7 & 87.2 & 64.0 & 96.1 & 75.6\\
   OC-SORT~\cite{cao2022observation}  & 82.0& 98.6 &  67.9 & 98.3  & 76.3\\
   \textbf{C-BIoU Tracker}  & \textcolor{magenta}{\textbf{89.2}} & \textbf{99.4} & \textbf{80.0} & \textbf{99.4} & \textbf{86.1}\\
   \hline
   \multicolumn{6}{l}{GMOT-40 Test Set~\cite{bai2021gmot}. Using Oracle Detections. }   \\
   DeepSORT~\cite{DeepSORT}  &86.4 & 87.9 & 84.9 & 94.2 & 88.6\\
   SORT~\cite{SORT}  &87.8 & 90.9 & 84.8 & 97.6 & 89.6\\
   OC-SORT~\cite{cao2022observation}  & 92.4& 99.3 &  86.0 & 98.5  & 90.0\\
   \textbf{C-BIoU Tracker}  & \textcolor{magenta}{\textbf{96.4}} & \textbf{99.7} & \textbf{93.2} & \textbf{99.6} & \textbf{95.6}\\
   \bottomrule
   \end{tabular}
   \label{table:comparisons_on_oracle_detection}
\end{table}

\begin{figure*}[h!]
   \centering
   \includegraphics[width=0.82\linewidth]{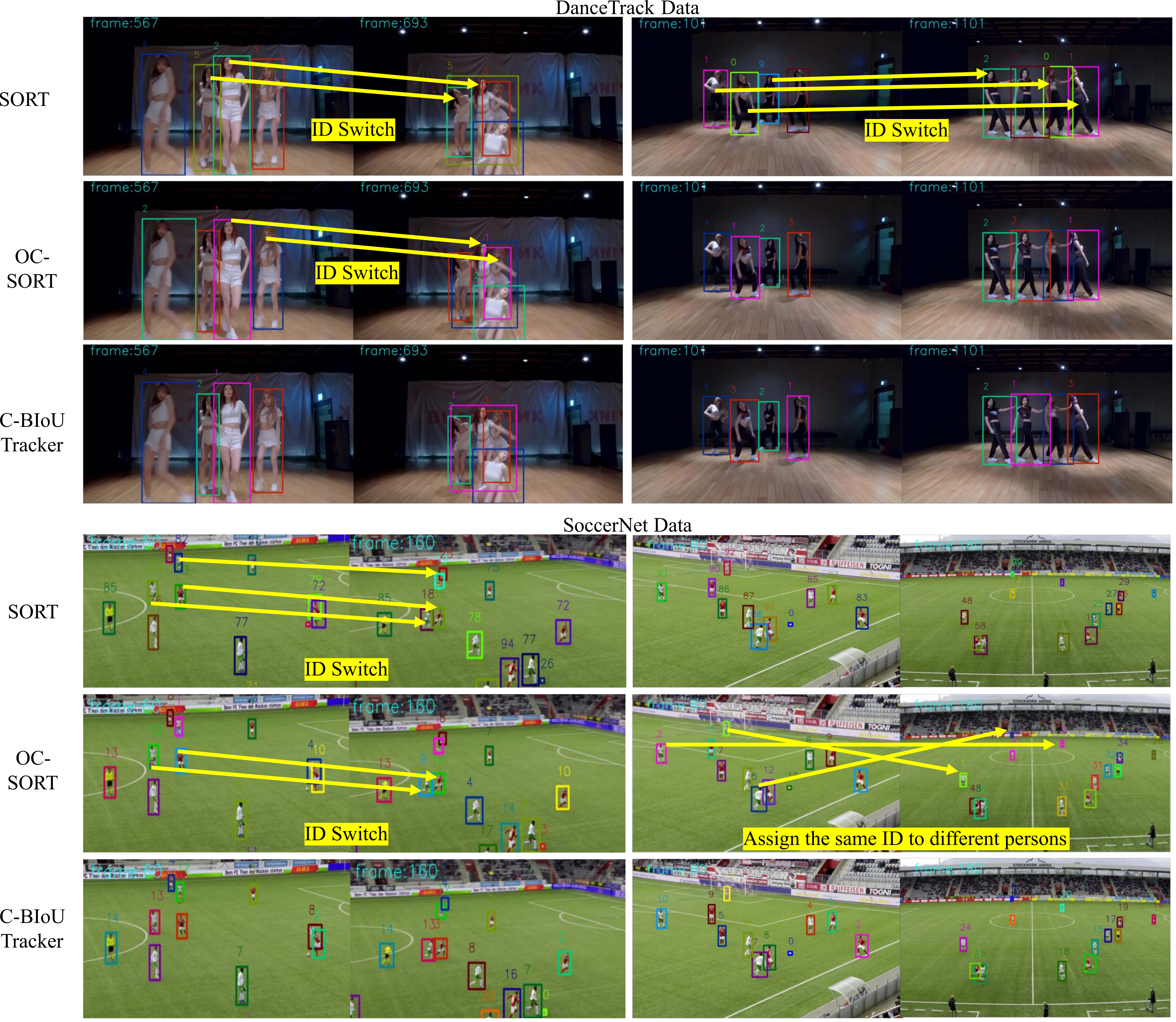}
   \caption{\textbf{Example results on the DanceTrack validation set~\cite{sun2022dancetrack} and SoccerNet test set~\cite{cioppa2022soccernet,deliege2021soccernet}.} Our C-BIoU tracker generates fewer tracking errors than SORT~\cite{SORT} and OC-SORT~\cite{cao2022observation}. }
   \label{fig:vis}
\end{figure*}

\subsubsection{Comparisons Using Estimated Detections}

Table~\ref{table:mot17_dancetrack_test} compares our C-BIoU tracker to mainstream MOT methods on the test sets of MOT17~\cite{MOT16} (private detections) and DanceTrack~\cite{sun2022dancetrack}. Each score is either from previous studies (\eg, DanceTrack~\cite{sun2022dancetrack}) or obtained by submitting the corresponding results to official evaluation servers. Note that, since the detection quality can significantly affect the overall tracking performance, for a fair comparison, methods in the bottom block use the detections generated by YOLOX~\cite{ge2021yolox}. The YOLOX weights for the MOT17 and DanceTrack datasets are offered by ByteTrack~\cite{sun2022dancetrack} and OC-SORT~\cite{cao2022observation}, respectively.
As methods in the top block may utilize better or worse detections than ours, we list them here for reference only. 

On the MOT17 test set, our method has a similar HOTA score as other methods. As analyzed in previous work~\cite{sun2022dancetrack}, the main bottleneck in MOT17 is detection other than tracking. On the DanceTrack test set, our method increases the HOTA score by a remarkable margin as compared to other methods. Although DeepSORT~\cite{DeepSORT}, SORT~\cite{SORT}, and ByteTrack~\cite{bytetrack} can generate comparable results on the MOT17 test set, their tracking performance largely drops on the DanceTrack test set, where more complicated object movements and similar bounding box scales are included.
Compared to the second-best method (\ie, OC-SORT~\cite{cao2022observation}), which applies the IoU, GIoU,  or DIoU for its matching, our C-BIoU tracker has increased the HOTA score by $4.9$ to make the new state of the art. Through the above comparisons, we prove the effectiveness of our C-BIoU tracker on conventional MOT data (\ie, MOT17) and our target MOT data (\ie, DanceTrack) that covers complicated motions and indistinguishable appearance.

In addition to the HOTA gain, our C-BIoU tracker can increase the inference speed of tracking (w/o the detection part). Table~\ref{table:inference_speed} reports the inference speed on the test sets of MOT17~\cite{MOT16} and DanceTrack~\cite{sun2022dancetrack}. Our C-BIoU tracker leverages the average speed of recent frames other than Kalman filters for its motion estimation. Therefore, we reduce the computation cost for data format transformation and other calculations used in Kalman filters. On the MOT17 and DanceTrack datasets, our C-BIoU tracker almost doubles the speed of OC-SORT~\cite{cao2022observation} and is much faster than other trackers. These results reveal that our C-BIoU tracker is a practical solution for real-world applications.

\subsubsection{Comparisons Using Oracle Detections}

To focus only on the tracking, we perform experiments using oracle detections from the DanceTrack validation set~\cite{sun2022dancetrack}, SoccerNet test set~\cite{cioppa2022soccernet,deliege2021soccernet}, and GMOT-40 test set~\cite{bai2021gmot}. The results in Table~\ref{table:comparisons_on_oracle_detection} indicate that our C-BIoU tracker can significantly surpass the other methods~\cite{SORT, DeepSORT, bytetrack, sun2022dancetrack, cao2022observation}, improving the tracking performance ranging from $2.6$ to $7.2$ in terms of the HOTA score. To obtain a more comprehensive look at the tracking performance, we plot the tracking results on multiple datasets for SORT~\cite{SORT}, OC-SORT~\cite{cao2022observation}, and our C-BIoU tracker in Fig.~\ref{fig:vis}.

Although we achieve the best performance on the three datasets, our tracking results are still imperfect even using oracle detections. Therefore, in the current research, it is useful to construct baselines using oracle detections and focus on improving the data association performance. We hope our baselines can motivate related research.

\begin{table}[t]
   \caption{\textbf{Ablation experiments on the DanceTrack validation set~\cite{sun2022dancetrack}, SoccerNet test set~\cite{cioppa2022soccernet,deliege2021soccernet}, and GMOT-40 test set~\cite{bai2021gmot}}. Where ``C.M.''  and ``Mo.'' represent the cascaded matching and motion estimation, respectively. We remove the cascaded matching and motion estimation in Fig.~\ref{fig:C-BIoU_framework} to construct \colorbox{lavender}{a unified framework} for the IoU, GIoU~\cite{giou2019}, DIoU~\cite{zheng2020distance}, and BIoU. The best results obtained by tuning the parameters are reported. Our BIoU performs better than the GIoU and DIoU. Using the C-BIoU setting is better than that using the BIoU alone. The motion estimation contributes to better HOTA scores.}
   \centering
   \setlength{\tabcolsep}{0.7pt}
   \footnotesize
   \begin{tabular}{ l c c ccccc}
   \toprule
   Tracker  ~~& C.M. ~~~&Mo. ~~ & HOTA$\uparrow$ & DetA$\uparrow$ & AssA$\uparrow$ & MOTA$\uparrow$ & IDF1$\uparrow$\\
   \midrule
   \hline
   \multicolumn{8}{l}{DanceTrack Validation Set~\cite{sun2022dancetrack}. Using Oracle Detections. }   \\
   \rowcolor{lavender}
   IoU Tracker &\xmark & \xmark & 76.6 & 97.5 & 60.2 & 99.2 & 73.6\\
   \rowcolor{lavender}
   GIoU Tracker &\xmark & \xmark &77.1 &\textbf{97.6} & 60.9  & 99.2& 74.0\\
   \rowcolor{lavender}
   DIoU Tracker &\xmark & \xmark & 75.1 & 97.0 & 58.2 & 99.2 & 72.9\\
   \rowcolor{lavender}
   BIoU Tracker &\xmark & \xmark & 80.0 & 97.5 & 65.7 & \textbf{99.3} & 78.2\\
   C-BIoU Tracker &\cmark & \xmark & 80.2 & 97.5 & 65.9 & \textbf{99.3} & 79.3\\
   \textbf{C-BIoU Tracker} &\cmark & \cmark &  \textbf{81.7} & \textbf{97.6} & \textbf{68.4} & \textbf{99.3} & \textbf{80.5}\\
   \hline
   \multicolumn{8}{l}{SoccerNet Test Set~\cite{cioppa2022soccernet,deliege2021soccernet}. Using Oracle Detections. }   \\
   \rowcolor{lavender}
   IoU Tracker &\xmark & \xmark  & 81.9 & 99.4 & 67.5 & \textbf{99.8} & 75.7\\
   \rowcolor{lavender}
   GIoU Tracker &\xmark & \xmark  &79.8 & \textbf{99.7} & 63.8 & 97.8 & 73.4\\
   \rowcolor{lavender}
   DIoU Tracker &\xmark & \xmark  & 84.3 & \textbf{99.7} & 71.2 & 99.2 & 79.9\\
   \rowcolor{lavender}
   BIoU Tracker &\xmark & \xmark  & 87.7 & \textbf{97.7} & 77.1 & 99.4 & 83.0\\
   C-BIoU Tracker &\cmark & \xmark & 88.9 & 99.5 & 79.4 & 99.5 & 85.2\\
   \textbf{C-BIoU Tracker} &\cmark & \cmark & \textbf{89.2} & 99.4 & \textbf{80.0} & 99.4 & \textbf{86.1}\\
   \hline
   \multicolumn{8}{l}{GMOT-40 Test Set~\cite{bai2021gmot}. Using Oracle Detections. }   \\
   \rowcolor{lavender}
   IoU Tracker &\xmark & \xmark  &93.0 & 99.6 & 86.8 & 98.1 & 90.1\\
   \rowcolor{lavender}
   GIoU Tracker &\xmark & \xmark  &93.4 & \textbf{99.8} & 87.4 & 98.5 & 90.2\\
   \rowcolor{lavender}
   DIoU Tracker &\xmark & \xmark  & 93.6 & 99.7 & 87.8 & 99.2 & 91.7\\
   \rowcolor{lavender}
   BIoU Tracker &\xmark & \xmark  & 96.2& 99.5 & 93.0 & \textbf{99.6} & 95.4\\
   C-BIoU Tracker &\cmark & \xmark   & 96.3 &  99.7 & 93.1 & \textbf{99.6} & 95.5\\
   \textbf{C-BIoU Tracker} &\cmark & \cmark & \textbf{96.4} & 99.7 & \textbf{93.2} & \textbf{99.6} & \textbf{95.6}\\
   \bottomrule
   \end{tabular}
   \label{table:ablation_on_oracle_detection}
\end{table}

\subsection{Ablation Experiments}
\label{sec:ablation}

We perform ablation studies to investigate the effect of individual modules and buffer scales in our C-BIoU tracker, as well as the effect of noisy detections.

\subsubsection{Effect of Each Module in the C-BIoU Tracker}
\label{sec:ablation_module}

Table~\ref{table:ablation_on_oracle_detection} shows the influence of each module in our C-BIoU tracker. In detail, we present the following analysis.

\noindent \textbf{Effect of the BIoU.} As a comparison, we apply the BIoU matching only once and remove the motion estimation in Fig.~\ref{fig:C-BIoU_framework} to construct the BIoU tracker. Using the same framework, the tracker equipped with BIoU achieves a higher HOTA score than other trackers equipped with IoU, GIoU~\cite{giou2019}, or DIoU~\cite{zheng2020distance}. Although the GIoU and DIoU can incorporate non-overlapping boxes for geometric consistency measurement, they may not generate comparable results as our BIoU does.

\noindent \textbf{Effect of Integrating Cascaded Matching and the BIoU.}  On the DanceTrack and GMOT-40, integrating cascaded matching and BIoU can slightly improve the performance as compared to using BIoU alone, with a HOTA gain of $0.2$ and $0.1$, respectively. While on SoccerNet, the improvement from integrating cascaded matching and the BIoU is more significant, with a HOTA gain of $1.2$. In the SoccerNet dataset, since the non-stationary camera can add extremely fast motion to objects, the use of cascade matching is more robust in this case.

\noindent \textbf{Effect of the Motion Estimation.} According to the results, motion estimation plays an important role in our C-BIoU tracker. Since our BIoU can compensate the matching space for incorrect motion estimation, using a simple motion estimation (\ie, averaging previous motions) yields better HOTA scores than that without using motion estimation.

\subsubsection{Effect of Buffer Scales in the C-BIoU Tracker}

\begin{figure}[h!]
   \centering
   \includegraphics[width=0.8\linewidth]{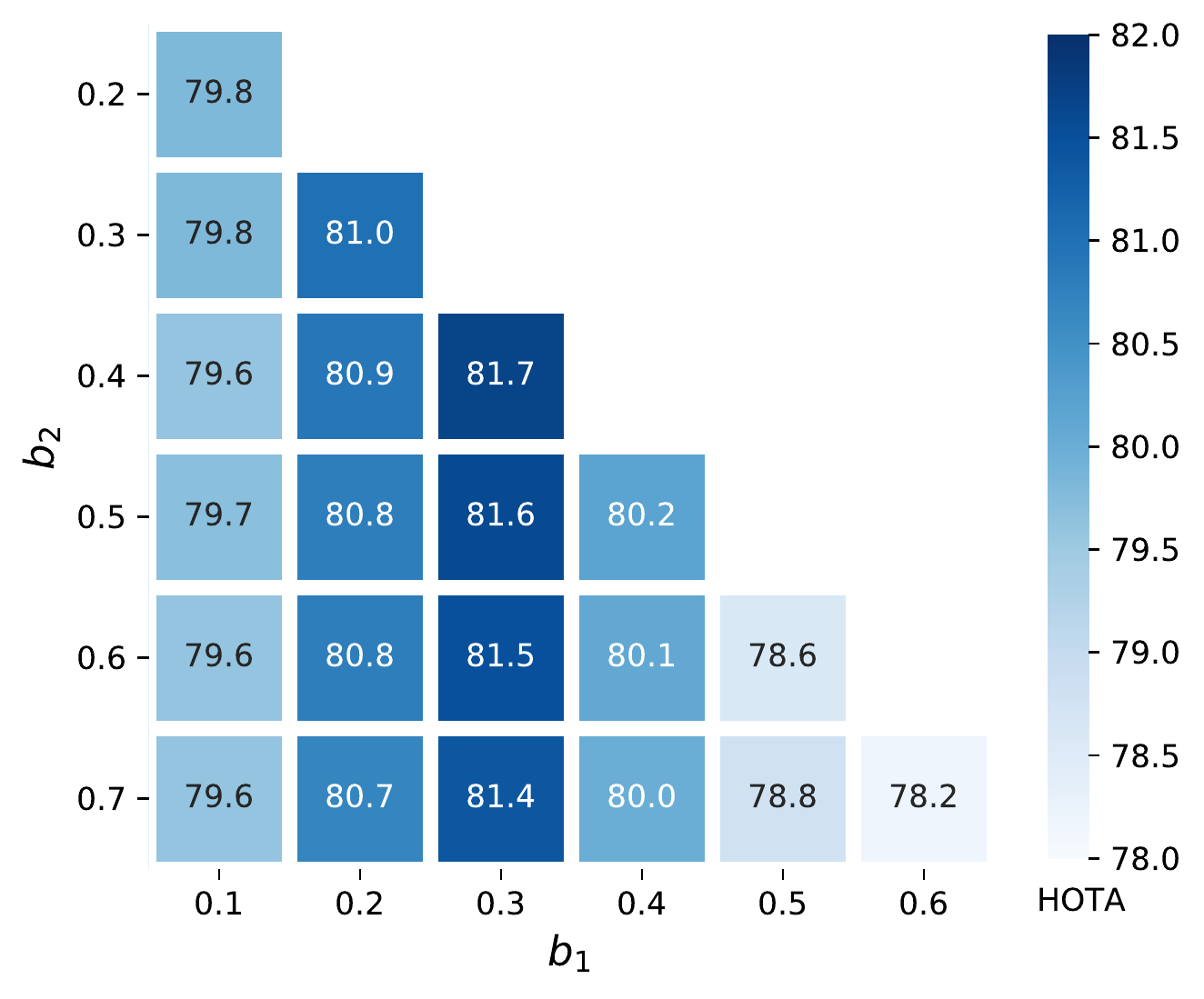}
   \caption{\textbf{Results of applying various buffer-scale combinations on the DanceTrack validation set~\cite{sun2022dancetrack}.} For buffer scales $b_{1}$ and $b_{2}$, since we have $b_{1}<b_{2}$, we only check the lower triangle of the combination matrix.}
   \label{fig:buffer_scales}
\end{figure}

In our C-BIoU tracker, the buffer scales $b_{1}$ and $b_{2}$ are critical hyperparameters. Here, we perform ablation studies to investigate how buffer scales affect the tracking performance. 
On the DanceTrack validation set~\cite{sun2022dancetrack}, we form the combination of $b_{1}$ and $b_{2}$ ranging from $0.1$ to $0.7$ and evaluate their tracking performance. Since we have $b_{1}<b_{2}$, we only need to check $21$ combinations. As shown in Fig.~\ref{fig:buffer_scales}, the combination of $[0.3, 0.4]$ gives the maximum HOTA score. In real practice, we perform a similar approach to select the best combination on the training dataset and apply them to the test dataset. Note that, although the variation of buffer scales affects the tracking performance remarkably, using the IoU tracker can only achieve a HOTA score of $76.6$, which is lower than using any of the above buffer combinations.

\subsubsection{Effect of the Detection Noise}
\label{sec:effect_detection_noise}

We have shown the superiority of our C-BIoU tracker in the previous experiments, however, we need to discuss about its limitations. Accordingly, we conduct the following analysis. 

In the previous experiments, our C-BIoU tracker significantly outperforms other MOT methods when using either high-quality detections generated by YOLOX~\cite{ge2021yolox} or oracle detections. Nonetheless, assuming that we only have low-quality detections, the robustness of our C-BIoU tracker needs to be studied. 
We inject noise (\ie, False Negatives and False Positives) to the oracle detections of the DanceTrack validation set~\cite{sun2022dancetrack} and form noisy detections that have quantitatively defined noise ratios. To inject detection noise, we first remove detections to generate False Negatives, and then add detections to non-target locations to form False Positives. Both of them have the same ratios. 

The results in Table~\ref{table:noise_study} reveal the influence of noisy detections on the tracking performance by considering noise ratios together. To date, such an ablation study had not been taken into account in existing studies. When the noise ratio is not higher than $20\%$, our C-BIoU tracker can maintain the best performance. However, a higher noise ratio, such as $40\%$, could lead to a worse performance of our C-BIoU tracker than the normal IoU tracker. The result is attributed to low-ratio noisy detections, which avoids the overlapping of the track and detection of an object in a \textit{short} interval of frames. Therefore, using BIoU matching to expand the matching space can result in more samples being correctly matched than IoU matching. However, for high-ratio noisy detections, the track and detection of an object do not overlap in a \textit{large} interval of frames. Consequently, both IoU matching and BIoU matching may generate tracking errors. In addition, the expansion of the matching space by BIoU leads to more aggressive matching, which increases the risk of missed matches with  False Positives. For these reasons, the robustness of our C-BIoU tracker decreases when extremely noisy detections are given. Fortunately, as reported in previous works~\cite{sun2022dancetrack, cioppa2022soccernet, bai2021gmot}, high-quality detections can be obtained in our target MOT datasets, since the similar appearance may ease the object detection. Thus, our C-BIoU tracker is applicable to real-world applications despite its limitations.

\begin{table}[t]
   \caption{\textbf{The influence of the detection quality.} We inject different levels of noises to the oracle detections of the DanceTrack validation set~\cite{sun2022dancetrack} to quantitatively investigate the influence of detection quality. IoU tracker and OC-SORT~\cite{cao2022observation} are used as baselines. We apply IoU matching only once in Fig.~\ref{fig:C-BIoU_framework} to construct the IoU tracker. }
   \centering
   \setlength{\tabcolsep}{.8pt}
   \footnotesize
   \begin{tabular}{ llccccc}
   \toprule
   Noise Ratio & Tracker & HOTA$\uparrow$ & DetA$\uparrow$ & AssA$\uparrow$ & MOTA$\uparrow$ & IDF1$\uparrow$\\
   \midrule
   \multirow{3}{*}{$0\%$ } 
   &OC-SORT~\cite{cao2022observation}  & 79.1& \textbf{97.7} &  64.0 & \textbf{99.6}  & 76.1\\
   & IoU Tracker & 76.6 & 97.5 & 60.2 & 99.2 & 73.6\\
   & \textbf{C-BIoU Tracker}  &  \textbf{81.7} &97.6 & \textbf{68.4} & 99.3 & \textbf{80.5}\\
   \hline
   \multirow{3}{*}{$20\%$ } 
   &OC-SORT~\cite{cao2022observation}  &61.4 &78.3 &48.1 &79.3 &65.3\\
   &IoU Tracker     & 57.6 & \textbf{79.5} & 41.7 & \textbf{81.7} & 59.6\\
   &\textbf{C-BIoU Tracker}  & \textbf{62.3} & 78.3 & \textbf{49.5} & 79.2 & \textbf{66.0}\\
   \hline
   \multirow{3}{*}{$40\%$ } 
   &OC-SORT~\cite{cao2022observation}  &28.0 &40.4 &19.4 &41.4 &34.3\\
   & IoU Tracker      & \textbf{38.3} & \textbf{58.6} & \textbf{25.0} & \textbf{60.4} & \textbf{40.8}\\
   & \textbf{C-BIoU Tracker}   & 29.2 & 58.0 & 14.7 & 57.7 & 29.1\\
   \bottomrule
   \end{tabular}
   \label{table:noise_study}
\end{table}

\section{Conclusion and Limitation Discussion}
\vspace{-1mm}
We present a novel Cascaded-Buffered IoU (C-BIoU) tracker to track multiple objects that have indistinguishable appearances and irregular motions. Experiments are conducted on related MOT datasets, and our C-BIoU tracker outperforms most existing methods by a notable margin. These results suggest that our C-BIoU tracker is generalizable and promising for tracking multiple objects with indistinguishable appearances and irregular motions.  The good performance of our C-BIoU tracker can be attributed to its buffered matching space, which mitigates the effect of irregular motions in two aspects: one is to directly match identical but non-overlapping detections and tracks in adjacent frames, and the other is to compensate for the motion estimation bias in the matching space.

As a limitation, our C-BIoU tracker may not be robust to extremely noisy detections (Sec.~\ref{sec:effect_detection_noise}). However, with advancements in object detection,  existing studies hint that good detections can be obtained in most MOT tasks. In addition, for other applications such as semi-automatic MOT annotations (\eg, \cite{fernandez2019semi,voigtlaender2019mots}), human factors are introduced to correct detections before tracking. 
Hence, our C-BIoU tracker remains a capable solution for real-world applications due to its simplicity, fast speed, and good tracking performance.

{\small
\bibliographystyle{ieee_fullname}
\bibliography{egbib}
}

\end{document}